\documentclass[pdflatex,sn-mathphys-num]{sn-jnl}% Math and Physical Sciences Numbered Reference Style
\usepackage{graphicx}%
\usepackage{multirow}%
\usepackage{amsmath,amssymb,amsfonts}%
\usepackage{amsthm}%
\usepackage{mathrsfs}%
\usepackage[title]{appendix}%
\usepackage{xcolor}%
\usepackage{textcomp}%
\usepackage{manyfoot}%
\usepackage{booktabs}%
\usepackage{algorithm}%
\usepackage{algorithmicx}%
\usepackage{algpseudocode}%
\usepackage{listings}%
\usepackage{longtable}% for long tables
\usepackage{multirow}
\usepackage{hhline}
\usepackage[load-configurations=version-1]{siunitx}
\usepackage{adjustbox}
\usepackage{float} % floatbarriers
\usepackage{listings}
\usepackage{hyperref}
\hypersetup{
    linkcolor=blue,
    citebordercolor=0.3 0.9 0.7,
    urlbordercolor=0.2 0.6 0.7,
    linkbordercolor=0.6 0.1 0.1,
}
\usepackage{enumitem}
\setlist[itemize]{
  label={\boldmath$\cdot$},
}
\theoremstyle{thmstyleone}%
\theoremstyle{thmstyletwo}%

\theoremstyle{thmstylethree}%

\begin{document}

\title[The CRISTAL Method]{The CRISTAL Method: Neurosymbolic analysis from AI-synthesized world models}
% \title[The CRISTAL Method]{The CRISTAL Method: Fast, reliable analytical problem-solving with pre-synthesized grounded world models}


\author*[1,4]{\fnm{Rafael} \sur{Kaufmann}}\email{rk@primordia.ai}
\equalcont{These authors contributed equally to this work.}
\author*[1,2]{\fnm{Felix} \sur{Neubürger}}\email{neubuerger.felix@fh-swf.de}
\equalcont{These authors contributed equally to this work.}

\author*[1,2]{\fnm{Michael} \sur{Walters}}\email{walters.michael@fh-swf.de}
\equalcont{These authors contributed equally to this work.}

\author[1,2]{\fnm{Thomas} \sur{Kopinski}}\email{kopinski.thomas@fh-swf.de}
\author[3]{\fnm{Dimitrije} \sur{Marković}}\email{dimitrije.markovic@tu-dresden.de}
\affil*[1]{\orgname{GAIA Lab}}
\affil[2]{\orgname{South Westphalia University of Applied Sciences}, \orgaddress{\street{Lindenstr 53}, \city{Meschede}, \postcode{59872}, \country{Germany}}}
\affil[3]{\orgdiv{Department}, \orgname{Technical University Dresden}, \orgaddress{\street{Zellescher Weg}, \city{Dresden}, \postcode{01069}, \country{Germany}}}
\affil*[4]{\orgname{Primordia Co.}}


\abstract{This project introduces the CRISTAL Method (Coherent Reliable Intentional Synthesis of Truthful Analysis Logic), a neurosymbolic framework for automating complex analysis workflows, with fundamental investment analysis as a primary use case. This domain poses major challenges: high structural uncertainty, noisy and subjective data, tight attention budgets, and the need for justified, reproducible decisions. 
Human analysts often struggle in this domain due to cognitive biases and limitations, suggesting significant value in automation. But while LLM-based agents have been proposed as analytical aids, their limitations---poor numerical reasoning, unawareness of uncertainty, and lack of reproducibility---hinder their effectiveness in this context. CRISTAL addresses these gaps through a principled blend of statistical model synthesis, continuous learning, and active learning. Starting from a natural-language prior knowledge curriculum, CRISTAL builds a dynamic, interpretable probabilistic program that enables full Bayesian inference, including uncertainty quantification and budget-aware data acquisition. CRISTAL continually refines its world model during analysis, leveraging LLMs for code synthesis and learning. We validate CRISTAL on a novel benchmark of synthetic equities with rich financial and textual data. On a company classification task, CRISTAL achieves Bayes-optimal accuracy with just 5 examples and a 5-second budget, outperforming state-of-the-art LLMs that plateau around 40\% accuracy even with order-of-magnitude more input data and compute.}




\maketitle

\section{Introduction}
We motivate the necessity for our technology first through the field of investment analysis.
A complex and cognitively demanding process, investment analysis requires sifting through vast amounts of data, integrating financial models with qualitative assessments, and making high-stakes decisions under conditions of uncertainty. The challenge is made worse by structural uncertainty, where data is often noisy, ambiguous, and expensive to obtain. Moreover, the need for justified, reproducible rationales is important, as investment decisions must be rigorously defended to stakeholders. These factors create an environment where human cognitive biases and limitations can significantly impact outcomes, raising the question of whether systematic, automated approaches can surpass human performance.
The remarkable capabilities of Large Language Models (LLMs) has sparked curiosity in their use as tools or even analysts in their own right, yet they exhibit inherent limitations that hinder their effectiveness in high-stakes domains like investment analysis, including:
\begin{itemize}
\item \textbf{Weak Numerical Reasoning}: LLMs often struggle with tasks requiring precise numerical computations and quantitative reasoning. Their training on vast textual data does not inherently equip them with robust mathematical problem-solving skills, leading to inaccuracies in scenarios demanding exact calculations. Studies have shown that while LLMs can handle basic arithmetic, they encounter difficulties with more complex mathematical reasoning tasks. \citep{mirzadeh2024gsmsymbolicunderstandinglimitationsmathematical, chiang2024overreasoningredundantcalculationlarge, wu2024reasoningrecitingexploringcapabilities, akhtar2023exploringnumericalreasoningcapabilities}

\item \textbf{Unreliable Uncertainty Quantification}: LLMs are prone to generating confident but incorrect responses, a phenomenon known as ``hallucination''. This issue partially stems from their inability to accurately express uncertainty, which is critical for assessing the reliability of their outputs in decision-making processes. Research indicates that LLMs often fail to convey appropriate levels of confidence, making it challenging to distinguish between correct and erroneous responses. \citep{nafar2024reasoninguncertaintextgenerative, xiong2024llmsexpressuncertaintyempirical, Farquhar2024, Huang_2025, fadeeva2024factcheckingoutputlargelanguage}

\item \textbf{Lack of Reproducibility}: The stochastic nature of LLMs contributes to variability in their outputs, even when presented with identical inputs. This lack of consistency poses significant challenges for reproducibility, a cornerstone of reliable analysis and decision-making. Without reproducible results, validating and trusting the conclusions drawn from LLMs becomes problematic. The importance of uncertainty quantification in ensuring model reproducibility has been highlighted in various studies. \citep{Volodina2021, bender2021parrots, Birhane2023, kim2024generativeartificialintelligencereproducibility}
\end{itemize}

More generally, it has been shown that agents seeking to perform robustly across distributional shifts \textit{must} learn approximately correct causal world models with small regret bounds \citep{richens2024robustagentslearncausal}. While advocates of pure deep learning (DL) argue that further pursuit of scale and more advanced training techniques can lead to the emergence of such approximations from data alone \citep{ge2024worldgptempoweringllmmultimodal}, recent evidence \citep{vafa2024evaluatingworldmodelimplicit} and long-standing theoretical arguments \citep{pearl2018theoreticalimpedimentsmachinelearning} indicate instead that the world models implicitly learned by DL processes will \textbf{not}, in general, be correct in the above sense. This indicates instead that robust, generalizable intelligence requires more than pattern recognition: it demands principled world modeling. However, rather than taking this to imply an ongoing need for handcrafted modeling, we propose instead that fully automated construction of approximately correct world models is feasible, by leveraging the capabilities of LLMs---not as world model learners, but as \textbf{samplers of meaning constructions} \citep{wong2023wordmodelsworldmodels}.

\begin{figure}
    \centering
    \includegraphics[width=0.9\linewidth]{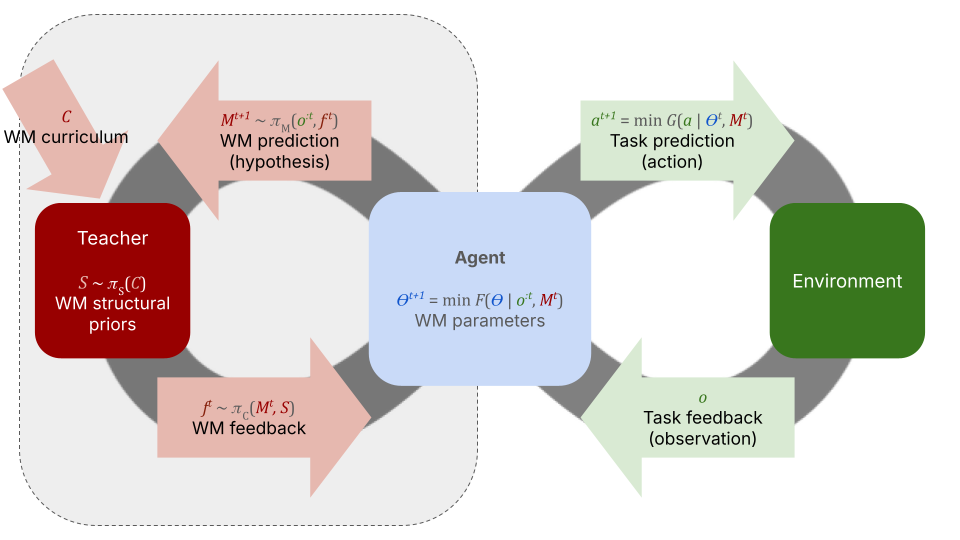}
    \caption{Schematic visualization of the CRISTAL Method. The Agent is acting as a predictor in the environment with an external world model and curriculum. The environmental feedback is calculated from the internal Teacher module teaching the world model and the Environment providing task feedback.}
    \label{fig:CRISTAL_pipeline}
\end{figure}

This summarizes our motivation for CRISTAL (Fig.\ \ref{fig:CRISTAL_pipeline}): a neurosymbolic framework designed to construct structured, probabilistic representations in complex, highly uncertain environments by integrating causal Bayesian reasoning into the core of automated analysis, enhancing the reliability and effectiveness of automated analysis workflows. CRISTAL is built upon a first-principles-driven methodology that integrates statistical model synthesis, continuous learning, and active learning. Unlike traditional machine learning approaches that rely solely on pattern recognition, CRISTAL constructs a probabilistic world model---a structured, interpretable representation of reality that serves as a foundation for decision-making. CRISTAL's model constructor module, CodeGen \citep{codegen2025}, automatically builds and refines world models as part of its automated continuous learning cycle. This module employs an LLM actor-critic validation loop evaluating model candidates that must be both approved by the critic LLM and validated against specified test cases.

The core of CRISTAL's methodology is its ability to synthesize an initial prior world model based on preexisting knowledge. In a financial context, this includes such materials as historical case studies, expert frameworks, and financial data. Crucially, the model is not static; it is designed as a probabilistic program, enabling the system to apply rigorous Bayesian inference techniques. These techniques allow CRISTAL to perform not only classification and prediction but also uncertainty quantification, model selection, and active learning---the strategic acquisition of new information based on compute budget constraints and data noise considerations.
A key feature of the CRISTAL approach is its alignment with the principle of progressive resolution. In real-world investment decision-making, analysts typically start with a rough, low-cost assessment of an opportunity before deciding whether to proceed with deeper analysis. Consequently, our method mimics this approach by efficiently determining whether an investment should be further examined and selecting the optimal strategy for additional information gathering. This efficiency ensures that decisions are made in a resource-conscious manner, balancing computational cost, data acquisition, and accuracy.
Further, CRISTAL implements continuous learning: it updates its world model parameters as new data is encountered. This process suggests the possibility of achieving asymptotically Bayes-optimal performance surpassing both human analysts and standalone LLMs in efficiency, accuracy, and robustness.

To demonstrate the effectiveness of our approach, we construct a benchmark of 200 synthetically-generated assets, each with a comprehensive set of financial and textual data, including historical financial statements, reports, and earnings call transcripts. In our study, we focus on a company classification sub-task, where CRISTAL achieves a Bayes-optimal accuracy of 100\% under true data-generating likelihoods, with a decision-making budget of just 5 seconds per task. In contrast, state-of-the-art LLMs achieve only 40\% accuracy, even when given ground-truth likelihoods and a larger budget of 20 seconds per task.
Thus, by integrating structured probabilistic reasoning with active and continuous learning, CRISTAL sets a new benchmark for intelligent decision-making under uncertainty. The implications extend beyond investment analysis, offering a potentially transformative paradigm for AI-driven decision support systems in domains where transparency, efficiency, and reproducibility are critical.

\section{Benchmark Dataset}
Financial forecasting relies heavily on high-quality datasets that accurately represent real-world market conditions. However, acquiring comprehensive financial data can be costly, subject to confidentiality constraints, and limited by historical biases. Further, crucially, in real-world financial situations it is almost always impossible to identify the true operating latent causes for any given outcome in the stock markets. Indeed, the very fact that a company's ``fundamental value'' is not directly measurable and is often drastically decoupled from its current market price in the short term, yet \textbf{must} influence its long-term stock performance, is both why fundamental investing is possible in the first place, and why it is a remarkably challenging discipline.

Hence, to illustrate our technique in a setting where ground-truth data is known, we introduce a benchmark dataset generation framework that creates realistic synthetic financial data for evaluating analytical models.
Our approach systematically generates synthetic stocks, categorized into different financial trajectories: \textit{high growth, stable}, and \textit{time-bomb}. 
Financial indicators are assigned through a structured sampling process, using probability matrices to ensure realistic variability. 
Hard financial indicators, such as revenue growth and leverage ratios, are generated using statistical distributions, while soft qualitative indicators are extracted from synthetically generated reports.
This dataset serves as the foundation for benchmarking CRISTAL against LLM-based analysts in financial forecasting. 
The models are tested on their ability to accurately classify assets and optimize decision-making under resource constraints. 
In this resource budgeting mechanism, models must self-regulate indicator selection based on LLM indicator extraction costs, simulating real-world financial analysis limitations.

\subsection{Indicator Generation}
The synthetic stock generation process consists of multiple steps designed to create realistic financial data for a specified number of synthetic stocks. This approach ensures diversity in financial characteristics while maintaining consistency across generated data.  
\paragraph{Company Categorization:}  
Synthetic stocks are generated and defined according to one of three categories, with the following predefined probabilities: \textit{high growth} (30\%), \textit{stable} (50\%), and \textit{time-bomb} (20\%). These categories represent different financial trajectories and risk profiles.  
\paragraph{Indicator Color Sampling:}  
Financial indicators (hard/quantitative and soft/qualitative) are assigned performance-based colors (\textit{red, yellow, green}) using predefined likelihood matrices specific to each asset category. 
These matrices encode the probability distribution of each indicator's color value based on the stock's assigned category.  
The likelihood matrices for the generation of the hard and soft indicators are shown in \autoref{tab:financial_likelihood_matrix} and \autoref{tab:soft_likelihood_matrix} of Appendix \ref{apd:supp}.

\paragraph{Indicator Value Sampling:}  
Quantitative values for indicators are generated based on the assigned colors. Different statistical distributions are used to ensure realistic variability:  
\begin{itemize}
    \item \textbf{Growth rates} (e.g., revenue growth, EBITDA growth) follow a \textit{beta distribution}.  
    \item \textbf{Financial ratios} (e.g., working capital ratio, leverage) follow a \textit{log-normal distribution}.  
    \item \textbf{Other metrics} follow a \textit{uniform distribution}.  
\end{itemize}  

\paragraph{Financial Statement Generation:}  
The sampled indicator values are used to construct synthetic financial statements, ensuring internal consistency:  
\begin{itemize}
    \item \textbf{Derived indicators} (e.g., net profit margin, ROIC, PE ratio) are computed based on sampled quantitative indicators.  
    \item \textbf{Core financial documents} (income statement, balance sheet, and cash flow statement) are generated using these computed values.  
    \item \textbf{Ratio consistency} is enforced to maintain realistic relationships among financial metrics.  
\end{itemize}  

\subsection{Generation of Company Reports}
Reports were created for each soft indicator to mimic the process of an analyst reading and assigning an indicator classification.
For this we take the ground truth data describing assets that was initially generated and utilize an LLM to write a report reflecting the ground truth data. 
The LLM is explicitly prompted to not write the actual name of the indicator or its color into the report to avoid ``cheating'' the subsequent qualitative interpretation (Soft Indicator Extraction) in the later phase. This system prompt is the same for each indicator and is given in the supplementary material in Appendix \ref{prompt:systemprompt_generator}.
For each soft indicator we designed an example descriptor prompt for the soft indicator to be passed into the generator. The \emph{environmental risk} indicator is displayed below.
\begin{quote}
{\scriptsize
{\ttfamily
A measure of the company’s exposure to risks related to environmental issues, such as climate change, regulatory changes, or resource scarcity.
Examples include penalties for non-compliance with emissions regulations or increased costs due to resource shortages.

    - Red: High environmental impact/risk \\
    - Yellow: Working on improvements \\
    - Green: Low impact/strong practices 
}
}
\end{quote} \label{SI_prompt}
    
\subsection{Extraction of Soft Indicators}

Part of the CRISTAL Method is extracting qualitative information from financial reports through the use of an LLM. In this Soft Indicator Extraction (SIE) step, the LLM is instructed to analyze a given report for a given target indicator with the same indicator description as the report generator. We used the \texttt{pydantic} python library \citep{Colvin_Pydantic_2025} to ensure a consistent normalized evaluation output, which was the model's estimation of indicator color (red, yellow, or green) and an open-ended reason field in English where the model could explain its decision.
% The return value is ensured to be json formatted with the indicator name, a value : \{red, yellow, green\} and a reason that explains the decision reason
% The json confinement is realized with the pydantic library \citep{Colvin_Pydantic_2025} and the LLM has three tries to give a correctly formatted output. If the soft indicator extraction fails the value \emph{None} is assigned and a more detailed log of the reasons is written into the json object.
The system prompt for the soft indicator extraction is given in Appendix \ref{prompt:systemprompt_extractor}:
This soft indicator extractor is used for benchmarking the generation and extraction purposes and for the use in the CRISTAL and LLM evaluations of the assets.
We investigated and compared the quality of reports from a variety of LLMs in our extraction benchmarking study (Appendix \ref{apd:si_benchmark}), and selected the highest performing models for subsequent experiments.

\section{Methodology}

The CRISTAL method is designed to evaluate and select financial assets based on a combination of quantitative and qualitative indicators. 
One of the core programs is to update the asset category belief likelihood matrix through inferencing on the observed indicators. 
This process is also resource budget-conscious, ensuring that the inference remains within a predefined affordance.

\begin{itemize}

\item \textbf{Process Quantitative Indicators:} Quantitative/hard indicator values can be readily computed from the provided asset data. In our experiment, observation data for indicators like \textit{revenue} are passed directly into the function, since quantitative values would be trivial for LLMs to recover from documents. Since these incur no compute overhead as a result of LLM indicator extraction on reports data, these are processed first.

\item \textbf{Process Qualitative Indicators:}
\begin{itemize}
    \item \textbf{Update soft indicator utility}: Soft indicator extraction is costly, as the LLM must interpret financial reports to generate an indicator estimation. Thus, we include a basic ranking algorithm step where indicators are first ranked according to their expected Bayes factor before extraction. This stage could be tuned as needed, depending on context.

    \item \textbf{Soft indicator extraction}: Within whatever budget remains, SIE is performed across soft indicators in descending Bayes factor utility to estimate their color.
    
\end{itemize}

\item \textbf{Bayesian Updating:} Bayesian updating is performed to refine the asset category probabilities from the observed and extracted indicators. For our categorical model, we used a Dirichlet prior.

\item \textbf{Final Class Prediction:} Once the budget is exhausted or all indicators have been processed, we store the posterior belief of this asset's category. In our experiments, we select the highest probability category as the model's answer.

\end{itemize}

\subsection{Soft Indicator Generation \& Extraction} 
\label{sec:sie-benchmark}
There is inherent aleatoric noise in the soft indicator generation-extraction pipeline, going from color~$\rightarrow$~report (generation), and then report~$\rightarrow$~value (extraction). 
If extracted values do not match generator values, this places a limitation on the performance CRISTAL can achieve in its likelihood inferencing process and any other downstream tasks.
Since this is LLM model-dependent, we performed a cursory evaluative sweep across the open-source models DeepSeek-r1:14b, DeepSeek-r1:70b \citep{deepseekai2025deepseekr1incentivizingreasoningcapability} and Llama3:70b \citep{grattafiori2024llama3herdmodels} as generators, and DeepSeek-r1:1.5b, DeepSeek-r1:7b, DeepSeek-r1:14b and Llama3.1:8b as extractors. Note that we tested larger models for the generator process but smaller ones for the extractor process since extraction is part of the CRISTAL algorithm where resource efficiency is of interest. Conversely, generation is done beforehand, so we were interested to see what effect, if any, larger models may have in pushing up the generator-extraction fidelity limit.

To compare generator and extractor models we employ a standard multiclass classification evaluation. The ground truth indicator values could be one of the three rating colors, and same for the generator with the addition of a \texttt{null} class on failed extraction. 
We report the accuracy, class precision, recall, F1-score and Matthews correlation coefficient (MCC) results in Appendix \ref{apd:si_benchmark}.

\section{Results}
\label{sec:results}
In the following, we compare the CRISTAL Method to an LLM-baseline analyst subject to our benchmark financial forecasting tasks, probing different environmental variables to determine if and to what degree CRISTAL trumps LLM analysts.
% The evaluation focuses on three key aspects: classification accuracy, quantile estimation performance, and budgeting efficiency under resource constraints.
% We compare how well each model classifies synthetic stocks into their respective categories, predicts financial growth quantiles, and optimally selects indicators while adhering to extraction cost limitations. 
% Additionally, we assess the accuracy of the soft indicator extraction process by measuring its ability to reconstruct qualitative financial indicators from generated reports to estimate the uncertainty from the extraction process.

Regarding the indicator fidelity analysis (\S\ref{sec:sie-benchmark}), we found that even a small model like Llama3.1:8b is a capable soft indicator extractor with a mean accuracy of $94.6\%$ and a Matthews Correlation Coefficient (MCC) of $0.92$. Additionally, its accuracy is independent across the three generator models. See Appendix \ref{apd:si_benchmark} for full results. Thus, we selected this model for extraction and Llama3:70b generating reports.
Though the indicator fidelity is quite accurate, it is worth noting that ultimately, for the success of our overall study, perfect fidelity is not necessary as long as it is not the dominant limiting factor on both CRISTAL and the LLM-Baseline analysts.

\subsection{Asset Category Prediction: CRISTAL Method vs.\ LLM analyst}
\label{sec:gt-exp}
%needs updated tables and confusion matrices 

%%%%%%%%%%%%%%% CHECK THE NUMBERS %%%%%%%%%%%%%
The empirical analysis reveals a fundamental performance gap between the LLM and CRISTAL analysts. 
Both analysts were provided the ground-truth likelihood matrix that was used to generate the synthetic asset data. Being a statistical program, CRISTAL directly uses the likelihood matrix in its inferencing, and for the LLM analyst the matrix is (necessarily) provided through prompting. The aim is thus to compare performance even under ideal conditions where the analysts know the asset category probabilities given indicator observations.

CRISTAL achieves Bayes optimal classification accuracy under the 88\% informativity constraint (as described in Appendix \ref{apd:informativity}) of the ground-truth likelihood matrix, minimizing expected risk through explicit alignment with the true posterior distributions (\autoref{tab:classification_report_CRISTAL}). 
In contrast, the LLM analyst exhibits systematic suboptimality, reaching an accuracy of only 35\% (\autoref{tab:classification_report_llm}). 
We posit that the performance gap likely originates from the inability of LLMs to preserve task-specific feature-label mutual information and efficient budgeting logic. Another explanation could be the limited reasoning capabilities and lack of mathematical context for the task at hand.

\begin{table}[htbp]
\centering
\caption{Classification report for CRISTAL results. Accuracy: 0.88, MCC: 0.80}
\label{tab:classification_report_CRISTAL}
\vspace{1em}

\begin{tabular}{lrrrr}
\toprule
 & Precision & Recall & F1-score & Support \\
\midrule
High Growth & 0.87 & 0.91 & 0.89 & 64 \\
Stable & 0.89 & 0.88 & 0.88 & 98 \\
Time-bomb & 0.86 & 0.82 & 0.84 & 38 \\
\hline
Macro Avg & 0.87 & 0.87 & 0.87 & 200 \\
Weighted Avg & 0.88 & 0.88 & 0.87 & 200 \\
\bottomrule
\end{tabular}

\end{table}

\begin{table}[htbp]
\centering
\caption{Classification report for Llama3:70b LLM analyst. Accuracy: 0.35, MCC: 0.00}
\label{tab:classification_report_llm}
\vspace{1em}

\begin{tabular}{lrrrr}
\toprule
 & Precision & Recall & F1-score & Support \\
\midrule
High Growth & 0.33 & 0.67 & 0.44 & 64 \\
Stable & 0.47 & 0.22 & 0.30 & 98 \\
Time-bomb & 0.17 & 0.11 & 0.13 & 38 \\
\hline
Macro Avg & 0.32 & 0.33 & 0.29 & 200 \\
Weighted Avg & 0.37 & 0.34 & 0.32 & 200 \\
\bottomrule
\end{tabular}

\end{table}

% confusion matrices side by side with a minipage?
% add the other plots here and write some stuff about it

\subsection{Few-shot learning and likelihood inference}
\label{sec:nshot-exp}
% describe setup of second experiment with starting with a uniform prior. 
% then n examples of each class are shown to the moedls
% cristal uses a bayesian update function
% llm gets quasi n shots to figure it out
% show results there with increasing n and different models
Whereas the previous experiment provided the ground-truth likelihood matrices to both architectures, in practice this must be constructed and inferred from observations. In the following experiment, CRISTAL applies Bayesian inference on a uniform prior of the indicator likelihoods using $n = [1, 5, \dots ]$ observations before proceeding with the asset categorization algorithm.
Each asset observation contains its indicator values and asset category.
This likelihood inference function in CRISTAL is a critical component for estimating the probability distribution of asset categories based on observed indicators. 
For the LLM analyst counterpart, instead of the ground-truth matrix, we also apply prompting with $n$ example asset observations (including their category label) to provide it the opportunity to extract the underlying feature correlations.

\begin{figure}
    \centering
    \includegraphics[width=1\linewidth]{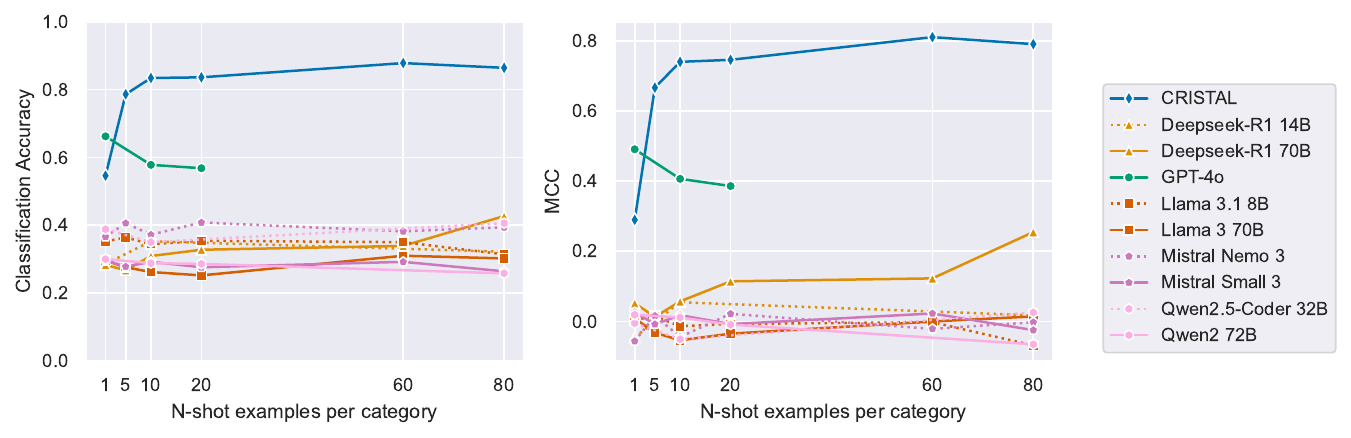}
    \vspace{-2em}
    \caption{Accuracy and MCC in n-shot learning context.}
    \label{fig:exp2-nshotplot}
\end{figure}

Both open-source (Llama, Deepseek, Mistral, Qwen2) and API models (GPT-4o, Gemini) were assayed in our evaluation. 
Once again, CRISTAL shows a dominating performance at asset classification, and a marked n-shot efficiency (\autoref{fig:exp2-nshotplot}). With $n=1$ observation, CRISTAL approaches 60\% accuracy (MCC=0.3), by $n=5$ performance already approaches that of the ground-truth likelihood configuration (\S\ref{sec:gt-exp}), and by $n=10$ begins to plateau at this maximum. 
Tabulated data can be found in Appendix \ref{apd:supp}, \autoref{tab:nshots}.

We find that with increasing number of examples shown to the LLMs the performance does not increase even when context size permits it, with the exception being the DeepSeek-R1-70b Model.
The pronounced trend among the LLM analysts is that of random guessing, indicated by scores around MCC=0. This is despite being provided with all the asset reports and ground-truth data.
Two outlying behaviors are displayed. First, GPT-4o is the only competitive model at first, but displays a downward or plateauing trend immediately. It is difficult to intuit why a downward trend would occur in this early range, however this could be simply an anomaly from insufficient statistics. Unfortunately, the context-window also limited attempts with larger $n$. 
The other anomaly, bucks the trend with a sudden performance increment at $n=80$. This could once again be an artifact of insufficient statistics, or it successfully found some genuine feature in the n-shot data that helped its evaluation. Even if the latter reason were true, its accuracy in the extreme is still completely on par with the worst performers and below 50\%, offering no reliable value.

\subsection{Prior LLM Knowledge}
For better or worse, LLMs come with preexisting biases learned from their training data. In the following experiment, we aimed to glean what extent these biases skewed LLM n-shot learning. For this, we swapped the \textit{high growth} and \textit{time-bomb} labels to simulate a ``counterintuitive'' world. 
At first blush, this may seem unimportant, but in practice one would not want an LLM analyst to be continuously normalizing their observations towards learned biases. With strong biases, an LLM may expect a high/red \textit{mismanagement} indicator implies a time-bomb asset, but with switched labels we task the LLM to forego this biased notion and listen only to the data.

\begin{figure}
    \centering
    \includegraphics[width=1\linewidth]{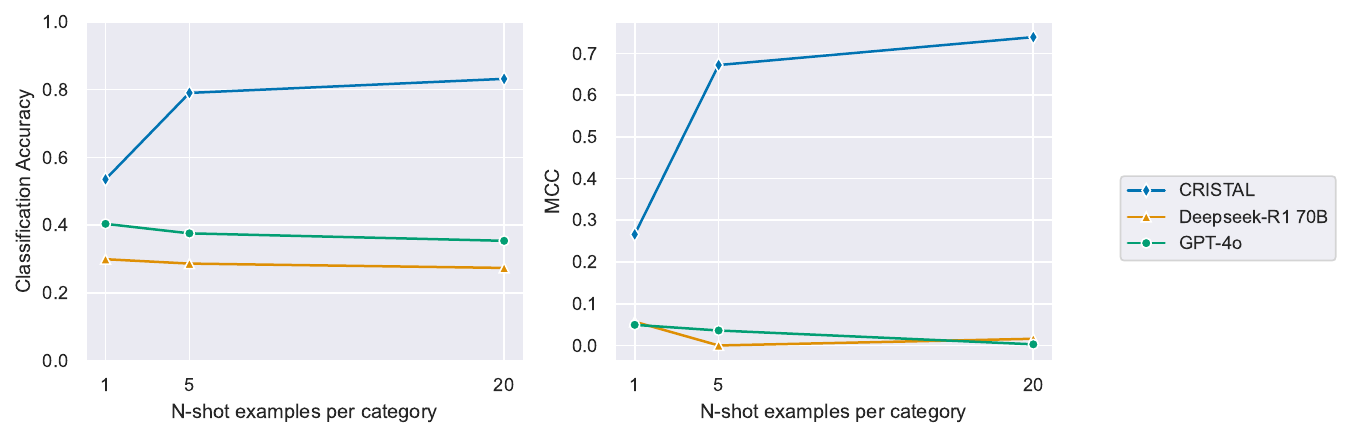}
    \vspace{-2em}
    \caption{Accuracy and MCC in n-shot learning context with swapped labels.}
    \label{fig:exp4-nshotplot}
\end{figure}

We targeted the two higher performing anomalous models from \S\ref{sec:nshot-exp} and repeat a portion of the n-shot experiment with swapped labels, shown in \autoref{fig:exp4-nshotplot}. We find that GPT-4o appears to lose its distinguishing edge, with its accuracy dropping to $\simeq 0.4$ on par with other models in \autoref{fig:exp2-nshotplot}, and no better than chance with $\text{MCC}\simeq 0$. The results are much the same for Deepseek-R1:70b, where presumably this counterintuitive labeling has reduced its performance to chance as well.
These results indeed suggest some measurable effect of these models holding onto their biases and ignoring what the data is telling them. Conversely, CRISTAL imparts no preconceived bias into the evaluation since it learns the likelihood matrix directly as the n-shot observations provide, and not through LLM ``black-box understanding''.

\section{Discussion}

Our experimental investigations consistently reinforce our thesis: even state-of-the-art LLM financial analysts struggle to holistically solve or ``understand'' tasks that rudimentary Bayesian analysis expeditiously can. The CRISTAL Method presents a neurosymbolic framework designed for this automated investment analysis under uncertainty, operating from a probabilistic world modeling approach. 
%combined with LLMs in precise, limited applications (i.e.\ indicator extraction) insulated from tasks that require numerical reasoning and uncertainty management.
% Our empirical findings are twofold. 
% First, in the soft indicator extraction benchmark, combining a high-capacity generative LLM (e.g., DeepSeek-r1:14b/70b) with a robust extractor (Llama3.1:8b) yields near-optimal performance for most indicators (up to 99.5\% accuracy, MCC 0.992). 
% However, some indicators (e.g., \emph{mismanagement}) remain challenging due to semantic ambiguity.
CRISTAL substantially outperforms pure LLM-based analysts on the asset classification task, achieving 88\% accuracy (MCC=0.80) using a probabilistic world model and active learning, approaching Bayes-optimality. 
LLMs, by contrast, reached only 35\% accuracy and exhibited poor mutual information retention and inference logic.
The few-shot learning experiment further highlights CRISTAL’s strength: it approaches optimal performance with just five examples using principled Bayesian updates, with a maximally uncertain uniform prior. 
LLMs, despite seeing the same data, perform near chance level, suggesting an inability to internalize feature-label relationships from limited context. 
Finally, the ``counterintuitive world'' label-swapping reinforces the problematic notion that LLM analysts bring biases from training data into their evaluations, instead of paying strict attention to the data.
The CRISTAL Method, using statistical inferencing algorithms with agnostic priors, is protected from such shortcomings. 
%The ``counterintuitive world'' test confirms this bias toward prior knowledge, reaffirming CRISTAL’s advantage in true data-driven inference.

\section{Future Research}
While CRISTAL shows strong promise, several open questions remain. Improving extraction for challenging indicators like \emph{mismanagement} will require better semantic modeling and domain adaptation. Future work will also validate the framework on real-world financial data beyond the current synthetic benchmark to assess robustness and practical utility.
Finally, deeper integration between LLMs and symbolic reasoning within CRISTAL could enhance adaptability. A more dynamic hybrid model may further improve efficiency, accuracy, and interpretability across diverse decision-making tasks.

\backmatter
\bibliography{sn-bibliography}% common bib file

\begin{appendices}

\section{LLM prompts} \label{apd:prompts}
This appendix contains the prompts used for generating synthetic reports and extracting soft indicators in the benchmarking process. These prompts were designed to simulate realistic financial analysis scenarios, guiding the models in interpreting qualitative data and making informed predictions. By providing structured and unstructured inputs, the prompts ensure consistency in evaluation while testing the adaptability of different AI approaches to financial forecasting tasks.
Prompt for the report generator:

{\scriptsize
\begin{quote} \label{prompt:systemprompt_generator}
{\ttfamily
You are a financial reporter writing a report on a company's
financial performance. You have been given the following information:

There are so called soft indicators that are used to evaluate 
the company's performance. The values of the indicator are
represented by the colors red, yellow, and green,
where red indicates a negative performance,
yellow indicates a neutral performance,
and green indicates a positive performance.
Your task is to write a business report
based on the given indicator and its value.

Do not use the actual indicator name anywhere the report.
Also do not use the actual value of the indicator in the report.
Do not give your report a headline that might give
away the indicator name. Format the report in markdown.
The description and value of the indicator are given below:

\{SOFT INDICATOR DESCRIPTION\}

The indicator value is: \{value\}
}
\end{quote}
}

Prompt for the soft indicator extractor
% also needs formatting
{\scriptsize
\begin{quote} \label{prompt:systemprompt_extractor} 
{\ttfamily
You are a financial analyst reading a report on a company's
financial performance. You have been given the following information:\\ 
There are so called soft indicators that are used to evaluate
the company's performance. The values of the indicator are represented
by the colors red, yellow, and green, where red indicates a negative
performance, yellow indicates a neutral performance, 
and green indicates a positive performance.\\
The indicator you are evaluating is: `\{indicator\}'
Your task is to read the business reports and infer the value
of the indicator. Your output should include your color estimate
(red, yellow, or green) for the given indicator.
Also give an explanation of why you chose that value.
The description of the indicator and the company reports are given below:

\{indicator description\}

<REPORTS>\\
\{report text\}\\
</REPORTS>

The indicator name to evaluate is: \{indicator\}
}
\end{quote}
}

\section{Soft indicator extraction benchmark} \label{apd:si_benchmark}

\begin{figure}[htbp]
    \centering
    \begin{minipage}{0.49\textwidth}
        \centering
        \includegraphics[width=\linewidth]{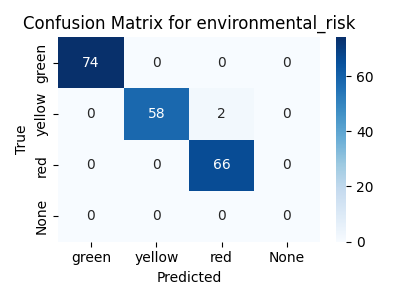}
        \caption{Confusion matrix for the soft indicator \emph{environmental risk} with the generator LLM: DeepSeek-r1:70b, extractor LLM: Llama3.1:8b}
        \label{fig:conf_environmental}
    \end{minipage}
    \hfill
    \begin{minipage}{0.49\textwidth}
        \centering
    \includegraphics[width=\linewidth]{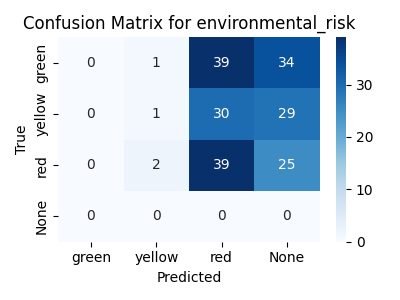}
    \caption{Confusion matrix for the soft indicator \emph{environmental risk} with the generator LLM: DeepSeek-r1:70b, extractor LLM: DeepSeek-r1:1.5b}
    \label{fig:conf_environmental_bad}
    \end{minipage}
\end{figure}

We benchmark our report generation and soft indicator extractor pipeline as described in \S \ref{sec:sie-benchmark}. 
An example for a benchmarking classification report is presented in \autoref{tab:SI_benchmark1}. As an additional visualization for evaluation a confusion matrix for each of the indicators is generated. An example for the soft indicator \emph{environmental risk} is shown in \autoref{fig:conf_environmental}
%mismanagement is weird DeepSeek14 is better here maybe we should try to explain?

\begin{table}[htbp]
\centering
\caption{Classification report for Soft Indicator extractor. Generator LLM : DeepSeek-r1:14b, extractor LLM: llama3.1:8b} \label{tab:SI_benchmark1}
    \begin{tabular}{lrrrrr}
        \toprule
        Indicator & Accuracy & Precision & Recall & F1 Score & MCC \\
        \midrule
        disruptor stagnant & 0.98 & 0.98 & 0.98 & 0.98 & 0.97 \\
        incumbent stagnant & 0.95 & 0.95 & 0.95 & 0.95 & 0.92 \\
        turnaround plan & 0.90 & 0.93 & 0.90 & 0.91 & 0.82 \\
        succession risk & 0.94 & 0.94 & 0.94 & 0.93 & 0.90 \\
        mismanagement & 0.67 & 0.78 & 0.67 & 0.67 & 0.56 \\
        geopolitical risk & 0.97 & 0.98 & 0.97 & 0.97 & 0.96 \\
        environmental risk & 0.99 & 0.99 & 0.99 & 0.99 & 0.98 \\
        product moat & 0.99 & 0.99 & 0.99 & 0.99 & 0.98 \\
        supply chain resilience & 0.97 & 0.97 & 0.97 & 0.97 & 0.95 \\
        regulatory pressure & 0.98 & 0.98 & 0.98 & 0.98 & 0.97 \\
        innovation pipeline & 0.98 & 0.98 & 0.98 & 0.98 & 0.98 \\
        customer concentration & 0.96 & 0.96 & 0.96 & 0.96 & 0.94 \\
        brand strength & 0.95 & 0.96 & 0.95 & 0.95 & 0.93 \\
        digital transformation & 0.99 & 0.99 & 0.99 & 0.99 & 0.99 \\
        \bottomrule
    \end{tabular}
\end{table}

When using the smaller LLM the extraction of soft indicators often times fail or are not accurate. This behaviour is shown in \autoref{fig:conf_environmental_bad}. It is not unexpected that smaller models can not comprehend the task given to them and subsequently fail the task and json confinement in addition to semantically not being able to solve the task.
More figures and tables can be found in the Appendix \ref{apd:supp} %supplementary material in the GitHub repository [redacted].
As a general rule we infer that the generation LLM with 70b parameters are both sufficient to generate comprehensive reports for the soft indicators. The DeepSeek-r1:14b model also generates the same quality in terms of this benchmark. Summarizing all combinations of models we calculate the mean and standard deviation over all indicator metrics. The results of the full benchmark are displayed in \autoref{tab:full_SI_benchmark_summary}
We observe that Llama3.1:8b performs best with DeepSeek-r1:14b in second place while DeepSeek-r1:7b and DeepSeek-r1:1.5b failing to get sufficient correct results. For performance reasons we choose Llama3.1:8b as our soft indicator extraction model in the other evaluations. This model is also the obvious choice when factoring in computational ressource costs since the DeepSeek-r1 models are larger and all containing a reasoning part in their output consuming more output tokens and thus requiring extra FLOPS. %more discussion on results?

\begin{tableorg}[htbp]
    \centering
    \caption{Summary of generator and extractor combination classification metrics sorted by MCC mean}
    \label{tab:full_SI_benchmark_summary}
\begin{adjustbox}{max width=\textwidth}
    \begin{tabular}{llrrrrrrrrrr}
        \toprule
         &  & \multicolumn{2}{c}{Accuracy} & \multicolumn{2}{c}{Precision} & \multicolumn{2}{c}{Recall} & \multicolumn{2}{c}{F1 Score} & \multicolumn{2}{c}{MCC} \\
         &  & mean & std & mean & std & mean & std & mean & std & mean & std \\
        generator model & extractor model &  &  &  &  &  &  &  &  &  &  \\
        \midrule
        deepseek-r1:14b & llama3.1:8b & 0.95 & 0.08 & 0.96 & 0.05 & 0.95 & 0.08 & 0.95 & 0.08 & 0.92 & 0.11 \\
        \cline{1-12}
        deepseek-r1:70b & llama3.1:8b & 0.95 & 0.08 & 0.96 & 0.05 & 0.95 & 0.08 & 0.95 & 0.08 & 0.92 & 0.11 \\
        \cline{1-12}
        llama3:70b & llama3.1:8b & 0.95 & 0.08 & 0.96 & 0.05 & 0.95 & 0.08 & 0.95 & 0.08 & 0.92 & 0.11 \\
        \cline{1-12}
        deepseek-r1:14b & deepseek-r1:14b & 0.94 & 0.07 & 0.96 & 0.04 & 0.94 & 0.07 & 0.94 & 0.06 & 0.92 & 0.09 \\
        \cline{1-12}
        deepseek-r1:70b & deepseek-r1:14b & 0.94 & 0.07 & 0.96 & 0.04 & 0.94 & 0.07 & 0.94 & 0.06 & 0.92 & 0.09 \\
        \cline{1-12}
        llama3:70b & deepseek-r1:14b & 0.94 & 0.07 & 0.96 & 0.04 & 0.94 & 0.07 & 0.94 & 0.07 & 0.92 & 0.09 \\
        \cline{1-12}
        deepseek-r1:70b & deepseek-r1:7b & 0.15 & 0.17 & 0.82 & 0.22 & 0.15 & 0.17 & 0.21 & 0.19 & 0.18 & 0.12 \\
        \cline{1-12}
        llama3:70b & deepseek-r1:7b & 0.15 & 0.16 & 0.81 & 0.18 & 0.15 & 0.16 & 0.21 & 0.19 & 0.18 & 0.12 \\
        \cline{1-12}
        deepseek-r1:14b & deepseek-r1:7b & 0.15 & 0.17 & 0.84 & 0.17 & 0.15 & 0.17 & 0.21 & 0.20 & 0.17 & 0.12 \\
        \cline{1-12}
        deepseek-r1:70b & deepseek-r1:1.5b & 0.32 & 0.09 & 0.32 & 0.15 & 0.32 & 0.09 & 0.24 & 0.06 & 0.01 & 0.03 \\
        \cline{1-12}
        llama3:70b & deepseek-r1:1.5b & 0.32 & 0.10 & 0.40 & 0.23 & 0.32 & 0.10 & 0.23 & 0.07 & 0.01 & 0.05 \\
        \cline{1-12}
        deepseek-r1:14b & deepseek-r1:1.5b & 0.32 & 0.09 & 0.33 & 0.22 & 0.32 & 0.09 & 0.23 & 0.07 & 0.01 & 0.05 \\
        \cline{1-12}
        \bottomrule
    \end{tabular}
\end{adjustbox}
\end{tableorg}

\section{Bayes optimal classification and informativity}
\label{apd:informativity}

Bayes Optimal Classification denotes the theoretically maximal classification performance attainable under a probabilistic framework, defined by minimizing the expected risk conditioned on the true posterior class probabilities and prior distributions of the data \cite{Duda2001}. This classifier serves as the statistical decision-theoretic benchmark, as no alternative algorithm---deterministic or stochastic---can achieve a lower misclassification rate under identical distributional assumptions \cite{Devroye1996}. When a method attains Bayes optimality, it implies that its decision boundaries align perfectly with the Bayes decision rule, given the modeled likelihoods and priors. Informativity (in our case calculated as 88\%) quantifies the relative reduction in class-label uncertainty conferred by the feature set, formalized via information-theoretic entropy measures. Specifically, it is defined as the ratio of the mutual information $I(Y; X)$ between features $ X $ and labels $ Y $ to the Shannon entropy $ H(Y) $ of the class distribution: 
\begin{equation*}
    \text{Informativity} = \frac{I(Y; X)}{H(Y)}.
\end{equation*}
An estimator achieving an accuracy value of 88\% indicates that the features resolve 88\% of the entropy (i.e., irreducible uncertainty) in $ Y $, with the remaining 12\% reflecting ambiguity inherent to the data-generating process \cite{Cover2005}. A Bayes optimal performance under this constraint demonstrates that it saturates the Cramér-Rao lower bound for classification error, achieving minimal possible risk given the 88\% informativity of the feature-conditioned likelihood matrix \cite{Kay1993}. This establishes that the classifier is statistically efficient, extracting all discriminative information available in the feature space.

\section{Benchmark dataset likelihood matrices} \label{apd:likelihood_matrices}
This appendix contains detailed tables outlining the indicator likelihoods used in the synthetic stock generation process and the soft indicator extraction costs applied in the budgeting task. The likelihood tables define the probability distributions of financial and qualitative indicators across different company categories, ensuring realistic data generation. The extraction cost tables specify the computational expense associated with retrieving soft indicators from textual reports, highlighting the resource trade-offs considered in model budgeting and decision-making.

\begin{table}[htbp]
\centering
\begin{tabular}{|l|l|l|l|}
\hline
\textbf{Category} & \textbf{Indicator} & \textbf{Red} & \textbf{Green} \\ \hline
\multirow{8}{*}{High Growth} & Gross Margin & 0.1 & 0.6 \\ \cline{2-4} 
 & Operating Margin & 0.3 & 0.4 \\ \cline{2-4} 
 & Asset Turnover & 0.2 & 0.4 \\ \cline{2-4} 
 & Leverage & 0.3 & 0.4 \\ \cline{2-4} 
 & Revenue Growth & 0.05 & 0.9 \\ \cline{2-4} 
 & EBITDA Growth & 0.1 & 0.6 \\ \cline{2-4} 
 & Working Capital Ratio & 0.4 & 0.3 \\ \cline{2-4} 
 & Reinvestment Rate & 0.1 & 0.8 \\ \hline
\multirow{8}{*}{Stable} & Gross Margin & 0.1 & 0.5 \\ \cline{2-4} 
 & Operating Margin & 0.1 & 0.5 \\ \cline{2-4} 
 & Asset Turnover & 0.1 & 0.4 \\ \cline{2-4} 
 & Leverage & 0.1 & 0.6 \\ \cline{2-4} 
 & Revenue Growth & 0.1 & 0.1 \\ \cline{2-4} 
 & EBITDA Growth & 0.1 & 0.3 \\ \cline{2-4} 
 & Working Capital Ratio & 0.1 & 0.5 \\ \cline{2-4} 
 & Reinvestment Rate & 0.2 & 0.1 \\ \hline
\multirow{8}{*}{Time-Bomb} & Gross Margin & 0.5 & 0.1 \\ \cline{2-4} 
 & Operating Margin & 0.6 & 0.1 \\ \cline{2-4} 
 & Asset Turnover & 0.6 & 0.1 \\ \cline{2-4} 
 & Leverage & 0.8 & 0.1 \\ \cline{2-4} 
 & Revenue Growth & 0.6 & 0.1 \\ \cline{2-4} 
 & EBITDA Growth & 0.6 & 0.1 \\ \cline{2-4} 
 & Working Capital Ratio & 0.7 & 0.1 \\ \cline{2-4} 
 & Reinvestment Rate & 0.7 & 0.1 \\ \hline
\end{tabular}
\caption{Financial Likelihood Matrix for the hard indicators in the benchmark dataset}
\label{tab:financial_likelihood_matrix}
\end{table}

\begin{table}[htbp]
\centering
\begin{tabular}{|l|l|l|l|}
\hline
\textbf{Category} & \textbf{Indicator} & \textbf{Red} & \textbf{Green} \\ \hline
\multirow{14}{*}{High Growth} & Disruptor Stagnant & 0.1 & 0.7 \\ \cline{2-4} 
 & Incumbent Stagnant & 0.6 & 0.2 \\ \cline{2-4} 
 & Turnaround Plan & 0.7 & 0.2 \\ \cline{2-4} 
 & Succession Risk & 0.3 & 0.4 \\ \cline{2-4} 
 & Mismanagement & 0.2 & 0.5 \\ \cline{2-4} 
 & Geopolitical Risk & 0.2 & 0.4 \\ \cline{2-4} 
 & Environmental Risk & 0.2 & 0.5 \\ \cline{2-4} 
 & Product Moat & 0.2 & 0.6 \\ \cline{2-4} 
 & Supply Chain Resilience & 0.4 & 0.3 \\ \cline{2-4} 
 & Regulatory Pressure & 0.4 & 0.3 \\ \cline{2-4} 
 & Innovation Pipeline & 0.1 & 0.7 \\ \cline{2-4} 
 & Customer Concentration & 0.3 & 0.4 \\ \cline{2-4} 
 & Brand Strength & 0.3 & 0.5 \\ \cline{2-4} 
 & Digital Transformation & 0.1 & 0.8 \\ \hline
\multirow{14}{*}{Stable} & Disruptor Stagnant & 0.6 & 0.2 \\ \cline{2-4} 
 & Incumbent Stagnant & 0.2 & 0.6 \\ \cline{2-4} 
 & Turnaround Plan & 0.7 & 0.2 \\ \cline{2-4} 
 & Succession Risk & 0.2 & 0.6 \\ \cline{2-4} 
 & Mismanagement & 0.2 & 0.6 \\ \cline{2-4} 
 & Geopolitical Risk & 0.2 & 0.5 \\ \cline{2-4} 
 & Environmental Risk & 0.3 & 0.4 \\ \cline{2-4} 
 & Product Moat & 0.2 & 0.6 \\ \cline{2-4} 
 & Supply Chain Resilience & 0.2 & 0.6 \\ \cline{2-4} 
 & Regulatory Pressure & 0.2 & 0.5 \\ \cline{2-4} 
 & Innovation Pipeline & 0.3 & 0.4 \\ \cline{2-4} 
 & Customer Concentration & 0.2 & 0.6 \\ \cline{2-4} 
 & Brand Strength & 0.2 & 0.6 \\ \cline{2-4} 
 & Digital Transformation & 0.3 & 0.4 \\ \hline
\multirow{14}{*}{Time-Bomb} & Disruptor Stagnant & 0.8 & 0.1 \\ \cline{2-4} 
 & Incumbent Stagnant & 0.3 & 0.5 \\ \cline{2-4} 
 & Turnaround Plan & 0.6 & 0.2 \\ \cline{2-4} 
 & Succession Risk & 0.7 & 0.1 \\ \cline{2-4} 
 & Mismanagement & 0.7 & 0.1 \\ \cline{2-4} 
 & Geopolitical Risk & 0.5 & 0.2 \\ \cline{2-4} 
 & Environmental Risk & 0.6 & 0.2 \\ \cline{2-4} 
 & Product Moat & 0.7 & 0.1 \\ \cline{2-4} 
 & Supply Chain Resilience & 0.6 & 0.2 \\ \cline{2-4} 
 & Regulatory Pressure & 0.6 & 0.2 \\ \cline{2-4} 
 & Innovation Pipeline & 0.7 & 0.1 \\ \cline{2-4} 
 & Customer Concentration & 0.6 & 0.2 \\ \cline{2-4} 
 & Brand Strength & 0.6 & 0.2 \\ \cline{2-4} 
 & Digital Transformation & 0.7 & 0.1 \\ \hline
\end{tabular}
\caption{Likelihood Matrix for the generation of soft indicators}
\label{tab:soft_likelihood_matrix}
\end{table}

\begin{table}[htbp]
\centering
\begin{tabular}{|l|c|}
\hline
\textbf{Indicator} & \textbf{Cost} \\ \hline
Disruptor Stagnant & 6 \\ \hline
Incumbent Stagnant & 3 \\ \hline
Turnaround Plan & 7 \\ \hline
Succession Risk & 9 \\ \hline
Mismanagement & 8 \\ \hline
Geopolitical Risk & 2 \\ \hline
Environmental Risk & 4 \\ \hline
Product Moat & 7 \\ \hline
Supply Chain Resilience & 8 \\ \hline
Regulatory Pressure & 5 \\ \hline
Innovation Pipeline & 6 \\ \hline
Customer Concentration & 7 \\ \hline
Brand Strength & 4 \\ \hline
Digital Transformation & 5 \\ \hline
\end{tabular}
\caption{Soft Indicator Costs for the budgeting task}
\label{tab:soft_indicator_costs}
\end{table}

\section{Supplementary tables and figures} \label{apd:supp}

This appendix presents additional figures and tables illustrating the benchmarking results for the soft indicators extraction benchmark and financial forecasting models. These figures provide a more detailed analysis of classification accuracy, quantile estimation performance, and budgeting efficiency across different scenarios. Visual comparisons highlight the strengths and limitations of each model, offering further insights into the impact of active learning versus static LLM-based approaches in financial analysis.
\begin{table}[htbp]
\centering
    \begin{tabular}{lrrrrr}
        \toprule
        Indicator & Accuracy & Precision & Recall & F1 Score & MCC \\
        \midrule
        disruptor stagnant & 0.29 & 0.10 & 0.29 & 0.15 & 0.01 \\
        incumbent stagnant & 0.37 & 0.19 & 0.37 & 0.25 & -0.06 \\
        turnaround plan & 0.17 & 0.76 & 0.17 & 0.12 & -0.00 \\
        succession risk & 0.29 & 0.78 & 0.29 & 0.28 & 0.09 \\
        mismanagement & 0.43 & 0.32 & 0.43 & 0.30 & 0.02 \\
        geopolitical risk & 0.27 & 0.34 & 0.27 & 0.22 & 0.01 \\
        environmental risk & 0.21 & 0.17 & 0.21 & 0.16 & 0.01 \\
        product moat & 0.49 & 0.29 & 0.49 & 0.35 & 0.05 \\
        supply chain resilience & 0.30 & 0.15 & 0.30 & 0.20 & 0.00 \\
        regulatory pressure & 0.24 & 0.15 & 0.24 & 0.19 & -0.08 \\
        innovation pipeline & 0.35 & 0.55 & 0.35 & 0.24 & -0.01 \\
        customer concentration & 0.38 & 0.18 & 0.38 & 0.24 & 0.01 \\
        brand strength & 0.38 & 0.30 & 0.38 & 0.27 & -0.05 \\
        digital transformation & 0.34 & 0.28 & 0.34 & 0.31 & 0.09 \\
        \bottomrule
    \end{tabular}
\caption{Classification report for Soft Indicator extractor. Generator LLM : DeepSeek-r1:14b, extractor LLM: DeepSeek-r1:1.5b}
\end{table}

\begin{table}[htbp]
\centering
    \begin{tabular}{lrrrrr}
        \toprule
        Indicator & Accuracy & Precision & Recall & F1 Score & MCC \\
        \midrule
        disruptor stagnant & 0.29 & 0.10 & 0.29 & 0.15 & 0.01 \\
        incumbent stagnant & 0.37 & 0.19 & 0.37 & 0.25 & -0.06 \\
        turnaround plan & 0.17 & 0.76 & 0.17 & 0.12 & -0.00 \\
        succession risk & 0.29 & 0.78 & 0.29 & 0.28 & 0.09 \\
        mismanagement & 0.43 & 0.32 & 0.43 & 0.30 & 0.02 \\
        geopolitical risk & 0.27 & 0.34 & 0.27 & 0.22 & 0.01 \\
        environmental risk & 0.21 & 0.17 & 0.21 & 0.16 & 0.01 \\
        product moat & 0.49 & 0.29 & 0.49 & 0.35 & 0.05 \\
        supply chain resilience & 0.30 & 0.15 & 0.30 & 0.20 & 0.00 \\
        regulatory pressure & 0.24 & 0.15 & 0.24 & 0.19 & -0.08 \\
        innovation pipeline & 0.35 & 0.55 & 0.35 & 0.24 & -0.01 \\
        customer concentration & 0.38 & 0.18 & 0.38 & 0.24 & 0.01 \\
        brand strength & 0.38 & 0.30 & 0.38 & 0.27 & -0.05 \\
        digital transformation & 0.34 & 0.28 & 0.34 & 0.31 & 0.09 \\
        \bottomrule
    \end{tabular}
\caption{Classification report for Soft Indicator extractor. Generator LLM : DeepSeek-r1:14b, extractor LLM: DeepSeek-r1:1.5b}
\end{table}

\begin{table}[htbp]
\centering
\begin{tabular}{lrrrr}
\toprule
 & Precision & Recall & F1-score & Support \\
\midrule
High Growth & 0.84 & 0.95 & 0.89 & 64 \\
Stable & 0.91 & 0.87 & 0.89 & 98 \\
Time-bomb & 0.91 & 0.82 & 0.86 & 38 \\
\midrule
Macro Avg & 0.89 & 0.88 & 0.88 & 200 \\
Weighted Avg & 0.89 & 0.89 & 0.88 & 200 \\
\bottomrule
\end{tabular}
\caption{Classification report for CRISTAL results, 5s. Accuracy: 0.89, MCC: 0.82}
\label{tab:classification_report_CRISTAL_5s}
\end{table}

\begin{table}[htbp]
\centering
\begin{tabular}{lrrrr}
\toprule
 & Precision & Recall & F1-score & Support \\
\midrule
High Growth & 0.78 & 0.97 & 0.87 & 64 \\
Stable & 0.87 & 0.82 & 0.84 & 98 \\
Time-bomb & 0.76 & 0.58 & 0.66 & 38 \\
\midrule
Macro Avg & 0.80 & 0.79 & 0.79 & 200 \\
Weighted Avg & 0.82 & 0.82 & 0.81 & 200 \\
\bottomrule
\end{tabular}
\caption{Classification report for CRISTAL results, 20s budget.~Accuracy:~0.82,~MCC:~0.71}
\label{tab:classification_report_CRISTAL_20s}
\end{table}

\begin{table}[htbp]
\centering
\begin{tabular}{lrrrr}
\toprule
 & Precision & Recall & F1-score & Support \\
\midrule
High Growth & 0.74 & 0.36 & 0.48 & 64 \\
Stable & 0.00 & 0.00 & 0.00 & 98 \\
Time-bomb & 0.22 & 0.97 & 0.36 & 38 \\
\midrule
Macro Avg & 0.32 & 0.44 & 0.28 & 200 \\
Weighted Avg & 0.28 & 0.30 & 0.22 & 200 \\
\bottomrule
\end{tabular}
\caption{Classification report for Llama3.1-8B. Accuracy: 0.30, MCC: 0.22}
\label{tab:classification_report_llama3:8b}
\end{table}

\begin{table}[htbp]
\centering
\begin{tabular}{lrrrr}
\toprule
 & Precision & Recall & F1-score & Support \\
\midrule
High Growth & 0.38 & 1.00 & 0.55 & 64 \\
Stable & 0.29 & 0.09 & 0.14 & 98 \\
Time-bomb & 0.00 & 0.00 & 0.00 & 38 \\
\midrule
Macro Avg & 0.22 & 0.36 & 0.23 & 200 \\
Weighted Avg & 0.26 & 0.36 & 0.24 & 200 \\
\bottomrule
\end{tabular}
\caption{Classification report for Llama3-70B. Accuracy: 0.36, MCC: 0.05}
\label{tab:classification_report_llama3:70b}
\end{table}

\begin{table}[htbp]
\centering
\begin{tabular}{lrrrr}
\toprule
 & Precision & Recall & F1-score & Support \\
\midrule
High Growth & 0.32 & 1.00 & 0.48 & 64 \\
Stable & 0.00 & 0.00 & 0.00 & 98 \\
Time-bomb & 0.00 & 0.00 & 0.00 & 38 \\
\midrule
Macro Avg & 0.11 & 0.33 & 0.16 & 200 \\
Weighted Avg & 0.10 & 0.32 & 0.16 & 200 \\
\bottomrule
\end{tabular}
\caption{Classification report for Qwen2-72B. Accuracy: 0.32, MCC: 0.00}
\label{tab:classification_report_Qwen2:72b}
\end{table}

\begin{table}[htbp]
\centering
\begin{tabular}{lrrrr}
\toprule
 & Precision & Recall & F1-score & Support \\
\midrule
High Growth & 0.41 & 0.88 & 0.56 & 57 \\
Stable & 0.75 & 0.09 & 0.15 & 70 \\
Time-bomb & 0.38 & 0.35 & 0.37 & 31 \\
\midrule
Macro Avg & 0.51 & 0.44 & 0.36 & 158 \\
Weighted Avg & 0.56 & 0.42 & 0.34 & 158 \\
\bottomrule
\end{tabular}
\caption{Classification report for DeepSeek-70B. Accuracy: 0.42, MCC: 0.18}
\label{tab:classification_report_DeppSeek:70b}
\end{table}

\begin{tableorg}[htbp]
\vspace*{-4.5em}  % Push content to center vertically
\centering
\begin{adjustbox}{max width=0.54\textwidth}
\begin{tabular}{|l|l|r|r|}
\hline
\textbf{N-shots} & \textbf{Model} & \textbf{Accuracy} & \textbf{MCC} \\ 
\noalign{\hrule height 2pt}
\multirow{10}{*}{1} & CRISTAL & 0.55 & 0.29 \\ \cline{2-4}
 & Deepseek-R1 14B & 0.29 & -0.05 \\ \cline{2-4}
 & Deepseek-R1 70B & 0.28 & 0.05 \\ \cline{2-4}
 & GPT-4o & \textbf{0.66} & \textbf{0.49} \\ \cline{2-4}
 & Llama 3 70B & 0.30 & 0.01 \\ \cline{2-4}
 & Llama 3.1 8B & 0.35 & 0.00 \\ \cline{2-4}
 & Mistral Nemo 3 & 0.37 & -0.06 \\ \cline{2-4}
 & Mistral Small 3 & 0.30 & 0.03 \\ \cline{2-4}
 & Qwen2 72B & 0.30 & 0.02 \\ \cline{2-4}
 & Qwen2.5-Coder 32B & 0.39 & -0.01 \\ \cline{2-4}
\noalign{\hrule height 2pt}
\multirow{6}{*}{5} & CRISTAL & \textbf{0.79} & \textbf{0.67} \\ \cline{2-4}
 & Deepseek-R1 70B & 0.27 & 0.01 \\ \cline{2-4}
 & Llama 3 70B & 0.28 & -0.03 \\ \cline{2-4}
 & Llama 3.1 8B & 0.36 & 0.01 \\ \cline{2-4}
 & Mistral Nemo 3 & 0.41 & 0.02 \\ \cline{2-4}
 & Mistral Small 3 & 0.28 & -0.01 \\ \cline{2-4}
\noalign{\hrule height 2pt}
\multirow{10}{*}{10} & CRISTAL & \textbf{0.83} & \textbf{0.74} \\ \cline{2-4}
 & Deepseek-R1 14B & 0.35 & 0.05 \\ \cline{2-4}
 & Deepseek-R1 70B & 0.31 & 0.06 \\ \cline{2-4}
 & GPT-4o & 0.58 & 0.41 \\ \cline{2-4}
 & Llama 3 70B & 0.26 & -0.05 \\ \cline{2-4}
 & Llama 3.1 8B & 0.34 & -0.01 \\ \cline{2-4}
 & Mistral Nemo 3 & 0.37 & -0.04 \\ \cline{2-4}
 & Mistral Small 3 & 0.29 & 0.02 \\ \cline{2-4}
 & Qwen2 72B & 0.29 & 0.01 \\ \cline{2-4}
 & Qwen2.5-Coder 32B & 0.35 & -0.05 \\ \cline{2-4}
\noalign{\hrule height 2pt}
\multirow{8}{*}{20} & CRISTAL & \textbf{0.84} & \textbf{0.75} \\ \cline{2-4}
 & Deepseek-R1 70B & 0.33 & 0.11 \\ \cline{2-4}
 & GPT-4o & 0.57 & 0.39 \\ \cline{2-4}
 & Llama 3 70B & 0.25 & -0.03 \\ \cline{2-4}
 & Llama 3.1 8B & 0.35 & -0.01 \\ \cline{2-4}
 & Mistral Nemo 3 & 0.41 & 0.02 \\ \cline{2-4}
 & Mistral Small 3 & 0.28 & -0.01 \\ \cline{2-4}
 & Qwen2 72B & 0.29 & -0.01 \\ \cline{2-4}
\noalign{\hrule height 2pt}
\multirow{6}{*}{60} & CRISTAL & \textbf{0.88} & \textbf{0.81} \\ \cline{2-4}
 & Deepseek-R1 70B & 0.34 & 0.12 \\ \cline{2-4}
 & Llama 3 70B & 0.31 & -0.00 \\ \cline{2-4}
 & Llama 3.1 8B & 0.35 & -0.00 \\ \cline{2-4}
 & Mistral Nemo 3 & 0.38 & -0.02 \\ \cline{2-4}
 & Mistral Small 3 & 0.29 & 0.02 \\ \cline{2-4}
\noalign{\hrule height 2pt}
\multirow{9}{*}{80} & CRISTAL & \textbf{0.86} & \textbf{0.79} \\ \cline{2-4}
 & Deepseek-R1 14B & 0.32 & 0.02 \\ \cline{2-4}
 & Deepseek-R1 70B & 0.43 & 0.25 \\ \cline{2-4}
 & Llama 3 70B & 0.30 & 0.01 \\ \cline{2-4}
 & Llama 3.1 8B & 0.31 & -0.07 \\ \cline{2-4}
 & Mistral Nemo 3 & 0.39 & -0.00 \\ \cline{2-4}
 & Mistral Small 3 & 0.26 & -0.02 \\ \cline{2-4}
 & Qwen2 72B & 0.26 & -0.07 \\ \cline{2-4}
 & Qwen2.5-Coder 32B & 0.41 & 0.03 \\ \cline{2-4}
\hline
\end{tabular}
\end{adjustbox}
\caption{Results Table for N-shot Analysis}
\label{tab:nshots}
\end{tableorg}
\end{appendices}

\end{document}